\documentclass[letterpaper,twocolumn]{IEEEtran}
\ifCLASSINFOpdf
  \usepackage[pdftex]{graphicx}
  \DeclareGraphicsExtensions{.pdf,.jpeg,.png}
\else
   \usepackage[dvips]{graphicx}
\fi
\usepackage[cmex10]{amsmath}
\usepackage{color}
\usepackage{amssymb}
\usepackage{bm}
\usepackage{caption}
\usepackage{lipsum}

\usepackage{float}
\usepackage{placeins}
\usepackage{multirow}
\usepackage{hhline}

\usepackage{algorithm,algorithmic}
\usepackage{array}
\ifCLASSOPTIONcompsoc
  \usepackage[caption=false,font=footnotesize,labelfont=sf,textfont=sf]{subfig}
\else
  \usepackage[caption=false,font=footnotesize]{subfig}
\fi
\usepackage{fixltx2e}
\usepackage{algorithm}
\usepackage{algorithmic}

\usepackage{amsthm}
\usepackage [autostyle, english = american]{csquotes}
\MakeOuterQuote{"}
\theoremstyle{plain}

\newcommand{\ds}{\displaystyle}

\newcommand{\De}{\Delta}

\newcommand{\mb}{\mathbb}
\newcommand{\mc}{\mathcal}

\newcommand{\tb}{\textbf}

\newcommand{\bea}{\begin{eqnarray}}
\newcommand{\eea}{\end{eqnarray}}
\newcommand{\beq}{\begin{equation}}
\newcommand{\eeq}{\end{equation}}

\newtheorem{ex}{Example}\newcommand{\Ex}{\begin{ex}\rm}
\newcommand{\eex}{\end{ex}}
\begin{document}
\title{Matrix Product State for Higher-Order Tensor Compression and Classification}
\author{Johann A. Bengua$^1$, Ho N. Phien$^1$, Hoang D. Tuan$^1$\thanks{$^1$Faculty of Engineering and Information Technology,  University of Technology Sydney, Ultimo, NSW 2007, Australia; Email: johann.a.bengua@student.uts.edu.au, ngocphien.ho@uts.edu.au, tuan.hoang@uts.edu.au.} and Minh N. Do$^2$\thanks{
$^2$Department of Electrical and Computer Engineering and the Coordinated Science Laboratory, University of Illinois at Urbana- Champaign, Urbana, IL 61801 USA; Email: minhdo@illinois.edu }}

\maketitle

\vspace*{-1.0cm}

\begin{abstract}
This paper introduces matrix product state (MPS) decomposition as a {\color{black}new and systematic method to}
compress  multidimensional data represented by higher-order tensors. It solves  two major bottlenecks
in tensor compression: computation and compression quality.  Regardless of tensor order, MPS compresses
tensors to \emph{matrices} of moderate dimension
which can be used for classification. Mainly based on a successive sequence of singular value decompositions (SVD), MPS is quite simple to implement {\color{black}and arrives at the global optimal matrix}, {\color{black}bypassing local alternating
optimization, which is not only computationally expensive but cannot yield the global solution}. Benchmark results show that MPS can  achieve better classification performance with favorable computation cost compared to other tensor compression methods.
\end{abstract}


\section{Introduction}\label{sec:introduction}

\IEEEPARstart{T}{here} is an increasing need to handle large multidimensional datasets that cannot efficiently be analyzed or processed using modern day computers.
Due to the curse of dimensionality it is urgent to develop mathematical tools which can evaluate information beyond the properties of large matrices \cite{Lu20111540}.
The essential goal is to reduce the dimensionality of multidimensional data, represented by tensors, with a minimal information loss by {\color{black}compressing} the original
tensor space to a lower-dimensional tensor space{\color{black}, also called the feature space} \cite{Lu20111540}.
Tensor decomposition is the most natural tool to {\color{black} enable} such compressions \cite{KOL09}.

Until recently, {\color{black} tensor compression is merely based on}  Tucker decomposition (TD) \cite{TUCKER1966}, also known as higher-order singular
value decomposition (HOSVD) when
orthogonality constraints on factor matrices  are imposed \cite{LAU00}.
TD is also an important tool for solving problems related to feature extraction, feature selection and classification of large-scale
multidimensional datasets in various research fields. Its well-known application in computer vision was introduced in \cite{VA02} to analyze some
ensembles of facial images represented by fifth-order tensors. In data mining, the HOSVD was also applied to identify handwritten digits \cite{SAV07}. In addition, the HOSVD has
been applied in neuroscience, pattern analysis, image classification and signal processing \cite{PHA10,KUA14,CIC15}.
The higher-order orthogonal iteration (HOOI) \cite{LAU00_1} is an alternating least squares (ALS)  for
finding the TD approximation  of a tensor. Its  application to independent component analysis (ICA) and simultaneous matrix diagonalization was investigated in \cite{DeLathauwer200431}. Another  TD-based method is multilinear principal component analysis (MPCA) \cite{LU08}, an extension of classical principal component analysis (PCA), which is closely related to HOOI.
{\color{black}Meanwhile, TD suffers the following conceptual bottlenecks in tensor compression:}
\begin{itemize}
\item {\it Computation.} TD  compresses an $N$th-order
tensor in tensor space $\mathbb{R}^{I_1\times I_2\times \cdots I_N}$ of large dimension $I=\prod_{j=1}^NI_j$
to its $N$th-order core tensor in a tensor space $\mathbb{R}^{\Delta_1\times \Delta_2\times\cdots \Delta_N}$ of smaller dimension $N_f=\prod_{j=1}^N\Delta_j$ by using $N$ factor matrices of size $I_j\times\Delta_j$.
Computation of these $N$ factor matrices is computationally intractable. Instead,
each factor matrix is alternatingly optimized with all other $N-1$ factor
 matrices held fixed, which is still computationally expensive. Practical application of the TD-based compression is normally limited to
 small-order tensors.
\item {\it Compression quality.}  TD is an effective representation of a tensor only when
the dimension of its core tensor is fairly large \cite{KOL09}. {\color{black}Restricting
dimension $N_f=\prod_{j=1}^N\Delta_j$ to a moderate size for tensor classification
results in significant lossy compression}, {\color{black}making TD-based compression a highly heuristic procedure
for classification.}
 It is also almost impossible to tune $\Delta_j\leq I_j$ among
$\prod_{j=1}^N\Delta_j\leq \bar{N}_f$ for a prescribed $\bar{N}_f$ to have a better compression.
\end{itemize}
In this paper, we introduce the matrix product state (MPS) decomposition \cite{VER04_1,VID03, VID04,PGA07} as a new method
to compress tensors, which fundamentally circumvent  all the above bottlenecks of TD-based compression. Namely,
\begin{itemize}
\item {\it Computation.} The MPS decomposition is fundamentally different from the TD in terms of its geometric structure as it is made up of local component tensors with maximum order three. Consequently, using the MPS decomposition for large higher-order tensors can potentially avoid the computational bottleneck of the TD and related algorithms.
{\color{black}Computation for orthogonal common factors in MPS  is based on successive SVDs without any recursive local optimization procedure and is very efficient with low-cost}.
\item {\it Compression quality.} MPS compresses $N$th-order tensors to their core matrices of size $\mathbb{R}^{N_1\times N_2}$. The {\color{black}dimension $N_f=N_1N_2$ can be easily tuned to a moderate size} with minimum information loss by pre-positioning the core matrix in the MPS decomposition.
\end{itemize}

MPS  has been proposed and applied to study quantum many-body systems with great success,
prior to its introduction to the mathematics community under the name tensor-train (TT) decomposition \cite{OSE11}.
However, to the best of our knowledge  its application to machine learning and pattern analysis
has not been proposed.

Our main contribution is summarized as follows:
\begin{itemize}
\item Propose MPS decomposition as a new {\color{black}and systematic} method for compressing tensors of arbitrary order to matrices of moderate dimension,
which circumvents  all existing bottlenecks in tensor compression;
\item Develop  MPS decomposition {\color{black}tailored for optimizing the dimensionality of the core matrices
and the compression quality.} {\color{black} Implementation issues of paramount importance for practical computation are discussed in detail. These include tensor mode permutation, tensor bond dimension control, and
    positioning the core matrix in MPS; }
\item Extensive experiments are performed along with comparisons to existing state-of-the-art tensor classification methods to show its advantage.
\end{itemize}
A preliminary result of this work was presented in \cite{Bengua_2015}. In the present paper, we rigorously introduce the MPS {\color{black} as a new and systematic} approach to tensor compression for classification, {\color{black}with computational complexity and efficiency analysis.} Furthermore, new datasets as well a new experimental design showcasing computational time and
{\color{black}classification success rate} (CSR) benchmarks are included.

The rest of the paper is structured as follows. {\color{black}Section \ref{Problemstatement} provides a rigorous mathematical analysis comparing MPS and TD in the context of tensor compression.
Section \ref{MPSalgorithms} is {\color{black}devoted to  MPS tailored for effective  tensor compression}, which
also includes a computational complexity analysis comparing MPS to HOOI, MPCA and uncorrelated multilinear discriminant analysis with regularization (R-UMLDA)  \cite{4711317}}. In Section \ref{Ressec}, experimental results are shown to benchmark all algorithms in classification performance and training time. Lastly, Section \ref{Conclusions} concludes the paper.

\section{MPS decomposition vs TD decomposition in tensor compression}\label{Problemstatement}
We introduce some notations and preliminaries of multilinear algebra \cite{KOL09}. Zero-order tensors are scalars and denoted by lowercase letters, e.g., $x$. A first-order tensor is a vector and denoted by boldface lowercase letters, e.g., \textbf{x}. A matrix is a second-order tensor and denoted by boldface capital letters, e.g., $\textbf{X}$. A higher-order tensor (tensors of order three and above) are denoted by boldface calligraphic letters, e.g.,
$\bm{\mc{X}}$.  Therefore, a general \textit{N}th-order tensor of size $I_1\times I_2\times\cdots\times I_N$ can
 be defined as $\bm{\mc{X}}\in\mathbb{R}^{I_1\times I_2\times\cdots\times I_N}$, where each $I_i$ is the dimension of its mode $i$. We also denote $x_i$, $x_{ij}$ and $x_{i_1\cdots i_N}$
as the $i$th entry $\textbf{x}(i)$, $(i,j)$th entry $\tb{X}(i,j)$ and $(i_1,\cdots,i_N)$th entry $\bm{\mc{X}}(i_1,\cdots,i_N)$  of  vector $\tb{x}$, matrix $\tb{X}$ and higher-order tensor $\bm{\mc{X}}$, respectively.

Mode-$n$ matricization (also known as mode-$n$ unfolding or flattening) of  $\bm{\mc{X}}$ is the process of unfolding or reshaping $\bm{\mc{X}}$ into a matrix $\tb{X}_{(n)}\in\mathbb{R}^{I_n\times (\prod_{i\neq n}I_i)}$ such that $\tb{X}_{(n)}(i_n,j)=\bm{\mc{X}}(i_1,\cdots,i_n,\cdots,i_N)$ for
$j=1+\sum_{k=1,k\neq n}^{N}(i_k-1)\prod_{m=1, m\neq n}^{k-1}I_m$. We also define the dimension of $\bm{\mc{X}}$ as
$\prod_{n=1}^NI_n$.
The mode-$n$ product of $\bm{\mc{X}}$ with  a matrix $\tb{A}\in\mathbb{R}^{J_n\times I_n}$ is denoted as $\bm{\mc{X}}\times_n \tb{A}$, which is
 a $N$th-order tensor of size $I_1\times\cdots\times I_{n-1}\times J_n\times I_{n+1}\times\cdots\times I_N$ such that
\[
\begin{array}{ll}
(\bm{\mc{X}}\times_n \tb{A})(i_1,\cdots, i_{n-1},j_n,i_{n+1},\cdots, i_N)&=\\
\ds\sum_{i_n=1}^{I_n}\bm{\mc{X}}(i_1,\cdots, i_n,\cdots i_N)\tb{A}(j_n,i_n).&
\end{array}
\]
The Frobenius norm of $\bm{\mc{X}}$ is defined as $
||\bm{\mc{X}}||_F=(\sum_{i_1=1}^{I_1}\sum_{i_2=1}^{I_2}\cdots\sum_{i_N=1}^{I_N}x^2_{i_1i_2\cdots i_{N}})^{1/2}$.

We are concerned with the following problem of {\color{black}tensor compression for supervised learning} :

{\it Based on
$K$ training \textit{N}th-order tensors $\bm{\mc{X}}^{(k)}\in\mathbb{R}^{I_1\times I_2\times\cdots\times I_N}$ ($k=1,2,\ldots,K$), find common factors to compress
both training tensor $\bm{\mc{X}}^{(k)}$ and test tensors $\bm{\mc{Y}}^{(\ell)}$ ($\ell=1,\cdots, L$) to a feature space of moderate dimension to enable  classification.}

Until now, only TD has been proposed  to address this problem \cite{PHA10}.
More specifically, the $K$ training sample tensors are firstly concatenated along the mode $(N+1)$ to form  an $(N+1)$th-order tensor $\bm{\mc{X}}$ as
\bea\label{tens1}
\bm{\mc{X}}=[\bm{\mc{X}}^{(1)} \bm{\mc{X}}^{(2)} \cdots \bm{\mc{X}}^{(K)} ]\in\mathbb{R}^{I_1\times I_2\times\cdots\times I_N\times K}.
\eea
TD-based compression  such as HOOI \cite{LAU00_1} is then applied  to have the approximation
\bea
\bm{\mc{X}}&\approx&\bm{\mc{R}}\times_1\tb{U}^{(1)}\times_2\tb{U}^{(2)}\cdots\times_N\tb{U}^{(N)},
\label{TD}
\eea
where each  matrix $\tb{U}^{(j)}\in\mathbb{R}^{I_j\times \Delta_j}$ $(j = 1,2,\ldots, N)$ is orthogonal, i.e. $\tb{U}^{(j)T}\tb{U}^{(j)}=\tb{I}$ ($\tb{I}\in\mathbb{R}^{\Delta_j\times \Delta_j}$ denotes the identity matrix). It is called a \textit{common factor} matrix and can be thought of as the principal components in each mode $j$. The parameters $\Delta_{j}$ satisfying
\begin{equation}\label{rank1}
\Delta_{j}\leq\text{rank}(\tb{X}_{(j)})
\end{equation}
are referred to as the compression ranks of the TD. \\
The $(N+1)$th-order core tensor $\bm{\mc{R}}$ and common factor matrices $\tb{U}^{(j)}\in\mathbb{R}^{I_j\times \Delta_j}$ are
{\color{black}supposed to be found} from the following nonlinear least squares
\begin{equation}\label{altopt}
\begin{array}{r}
\ds\min_{\substack{\bm{\mc{R}}\in\mathbb{R}^{\Delta_1\times\cdots\times \Delta_N\times K},\\ \tb{U}^{(j)}\in\mathbb{R}^{I_j\times\Delta_j}, j=1,...,N}}\Phi(\bm{\mc{R}},\tb{U}^{(1)},\cdots,\tb{U}^{(N)})\\
\mbox{subject to}\quad (\tb{U}^{(j)})^{T}\tb{U}^{(j)}=\tb{I}, j=1,...,N,\end{array}
\end{equation}
where $\Phi(\bm{\mc{R}},\tb{U}^{(1)},\cdots,\tb{U}^{(N)}):=||\bm{\mc{X}}-\bm{\mc{R}}\times_1\tb{U}^{(1)}\times_2\tb{U}^{(2)}\cdots\times_N\tb{U}^{(N)}||_F^2  $
The optimization problem (\ref{altopt}) is computationally intractable, which could be addressed only by
alternating least squares (ALS) in each $\tb{U}^{(j)}$ (with other $\tb{U}^{(\ell)}$, $\ell\neq j$ held fixed) \cite{LAU00_1}:{\color{black}
\begin{equation}\label{ad1}
\begin{array}{r}
\ds\min_{\substack{\bm{\mc{R}}^{(j)}\in\mathbb{R}^{\Delta_1\times\cdots\times \Delta_N\times K},\\ \tb{U}^{(j)}\in\mathbb{R}^{I_j\times\Delta_j}}}\Phi^{(j)}(\bm{\mc{R}}^{(j)},\tb{U}^{(j)})\\
\mbox{subject to}\quad (\tb{U}^{(j)})^{T}\tb{U}^{(j)}=\tb{I},\end{array}
\end{equation}
where $\Phi^{(j)}(\bm{\mc{R}}^{(j)},\tb{U}^{(j)}):=||\bm{\mc{X}}-\bm{\mc{R}}^{(j)}\times_1\tb{U}^{(1)}\times_2\tb{U}^{(2)}\cdots\times_N\tb{U}^{(N)}||_F^2 $.
}
The computation complexity per one iteration consisting  of $N$  ALS (\ref{ad1})  is \cite[p. 127]{Mariya11}
\begin{equation}\label{complex}
\mc{O}(K\De I^N + NKI\De^{2(N-1)}+NK\De^{3(N-1)})
\end{equation}
for
\begin{equation}\label{complex1}
I_j\equiv I\quad\mbox{and}\quad \Delta_j\equiv\Delta, j=1, 2,...,N.
\end{equation}
The optimal $(N+1)$th-order core tensor $\bm{\mc{R}}\in\mathbb{R}^{\Delta_1\times\cdots\times\Delta_N\times K}$
in (\ref{altopt}) is seen as the concatenation of {\color{black}compressed} $\widetilde{\bm{\mc{X}}}^{(k)}\in\mathbb{R}^{\Delta_1\times\cdots\times\Delta_N}$
of the sample tensors $\bm{\mc{X}}^{(k)}\in\mathbb{R}^{I_1\times\cdots\times I_N}$, $k=1,\cdots,K$:
\bea\label{tra1}
\bm{\mc{R}}=[\widetilde{\bm{\mc{X}}}^{(1)}\widetilde{\bm{\mc{X}}}^{(2)}\cdots\widetilde{\bm{\mc{X}}}^{(N)}]= \bm{\mc{X}}\times_1(\tb{U}^{(1)})^T\cdots\times_N(\tb{U}^{(N)})^T.
\eea
Accordingly, the test tensors $\bm{\mc{Y}}^{(\ell)}$ are compressed to
\bea\label{tens2}
\widetilde{\bm{\mc{Y}}}^{(\ell)}=\bm{\mc{Y}}^{(\ell)}\times_1(\tb{U}^{(1)})^T\cdots\times_N(\tb{U}^{(N)})^T
\in\mathbb{R}^{\Delta_1\times\cdots\times\Delta_N}.
\eea
The number
\bea
N_f=\prod_{j=1}^{N}\Delta_j
\label{NfTD}
\eea
thus represents the dimension of the feature space $\mathbb{R}^{\Delta_1\times\cdots\times\Delta_N}$.\\
{\color{black}Putting aside the computational intractability of the
optimal factor matrices  $\tb{U}^{(j)}$ in  (\ref{altopt}), the TD-based tensor compression by (\ref{tra1}) and
(\ref{tens2}) is {\color{black}a systematic procedure} only when
the right hand side of (\ref{TD}) provides {\color{black}a good approximation} of $\bm{\mc{X}}$, which
is impossible for small $\Delta_j$ satisfying (\ref{rank1}) \cite{KOL09}. In other words,
{\color{black}the compression of large dimensional tensors to small dimensional tensors results in substantial lossy compression
under the TD framework.} Furthermore, one can see
the value of  (\ref{ad1}) is lower bounded by
\begin{equation}\label{ad3}
\sum_{i=1}^{r_j-\Delta_j-1}s_i,
\end{equation}
where $r_j:=\text{rank}(\bm{\mc{X}}_{(j)})$ and  $\{s_{r_j},\cdots,s_1\}$ is
the set of non-zero eigenvalues of the positive definite matrix $\bm{\mc{X}}_{(j)}(\bm{\mc{X}}_{(j)})^T$ in decreasing order.
Since the matrix $\bm{\mc{X}}_{(j)} \in\mathbb{R}^{I_j\times(K\prod_{\ell\neq j}I_{\ell})}$ is {\color{black}highly unbalanced} as a result of tensor matricization along  one mode versus the rest, it is almost full-row (low) rank ($r_j\approx I_j$)
and its squared $\bm{\mc{X}}_{(j)}(\bm{\mc{X}}_{(j)})^T$ of size $I_j\times I_j$
is well-conditioned in the sense that its eigenvalues do not decay
quickly. As a consequence, (\ref{ad3}) cannot be small for small $\Delta_j$ so the ALS (\ref{ad1})
{\color{black}cannot result in a good approximation.}
The information loss with the least square (\ref{ad1})
is thus more than
\begin{equation}\label{tdloss1}
-\sum_{i=1}^{r_j-\Delta_j-1}\frac{s_i}{\sum_{i=1}^{r_j}s_i}
\log_2\frac{s_i}{\sum_{i=1}^{r_j}s_i},
\end{equation}
which is really essential in the von Neumann entropy \cite{NielsenChuang} of $\bm{\mc{X}}_{(j)}$:
\begin{equation}\label{tdloss2}
-\sum_{i=1}^{r_j}\frac{s_i}{\sum_{i=1}^{r_j}s_i}
\log_2\frac{s_i}{\sum_{i=1}^{r_j}s_i}.
\end{equation}
Note that each entropy (\ref{tdloss2}) quantifies only local correlation between mode $j$ and the rest \cite{BENGUA2016}.
The MPCA \cite{LU08} aims at (\ref{altopt}) with
\[
\bm{\mc{X}}=[(\bm{\mc{X}}^{(1)}-\bar{\bm{\mc{X}}})\cdots (\bm{\mc{X}}^{(K)}-\bar{\bm{\mc{X}}})]
\]
with $\bar{\bm{\mc{X}}}=\frac{1}{K+L}(\sum_{k=1}^K\bm{\mc{X}}^{(k)}+\sum_{\ell=1}^L\bm{\mc{Y}}^{\ell)})$.
With such definition of $\bm{\mc{X}}$, $(N+1)$th-order core tensor $\bm{\mc{X}}$ is the concatenation of
principal components of $\bm{\mc{X}}^{(k)}$, while principal components of $\bm{\mc{Y}}^{(\ell)}$
is defined by $(\bm{\mc{Y}}^{(\ell)}-\bar{\bm{\mc{X}}})\times_1(\bm{U}^{(1)})^T\cdots \times_N(\bm{U}^{(N)})^T$.
Thus, MPCA suffers the similar conceptual drawbacks
inherent by TD. Particularly, restricting $N_f=\prod_{j=1}^N\Delta_j$ {\color{black}to a moderate size leads to ignoring} many important principle
components.}

We now present a novel approach to extract tensor features, which is based on MPS. {\color{black}Firstly, \emph{permute}
all modes of the tensor $\bm{\mc{X}}$ and position mode $K$ such that
such that
\begin{equation}\label{pos}
\bm{\mc{X}}\in\mathbb{R}^{I_1\times\cdots I_{n-1}\times K\times I_{n}\cdots\times I_{N}},
\end{equation}
$I_1\geq\cdots\geq I_{n-1}$ and $I_{n}\leq\cdots...\leq I_N$. The elements of $\bm{\mc{X}}$ can be presented in the following}
\emph{mixed-canonical form} \cite{SCHO2011} of the matrix product
state (MPS) or tensor train (TT) decomposition \cite{PGA07,VID03,
VID04,OSE11}: {\color{black}
\bea x_{i_1\cdots k\cdots i_{N}}&=&x^{(k)}_{i_1\cdots
i_n\cdots i_{N}} \nonumber\\
&\approx&\tb{B}^{(1)}_{i_1}\cdots\tb{B}^{(n-1)}_{i_{n-1}}\tb{G}^{(n)}_{k}\tb{C}^{(n+1)}_{i_{n}}\cdots\tb{C}^{(N+1)}_{i_{N}},\nonumber\\
\label{mpsdef_mixed}
\eea}
where {\color{black}matrices}
$\tb{B}^{(j)}_{i_{j}}$ and $\tb{C}^{(j)}_{i_{(j-1)}}$ (the upper
index "$(j)$" denotes the position $j$ of the matrix in the chain)
of {\color{black}size} $\Delta_{j-1}\times\Delta_j$ ($\Delta_{0}=
\Delta_{N+1}=1$), are called "left" and "right" \textit{common
factors} which satisfy the following orthogonality conditions: \bea
\label{LeftCan1}
\sum_{i_j=1}^{I_j}(\tb{B}^{(j)}_{i_j})^{T}\tb{B}^{(j)}_{i_j}&=&\tb{I},\quad(j=1,\ldots,n-1)
\eea and \bea \label{RightCan1}
\sum_{i_{j-1}=1}^{I_{j-1}}\tb{C}^{(j)}_{i_{j-1}}(\tb{C}^{(j)}_{i_{j-1}})^{T}&=&\tb{I},~~(j=n+1,\ldots,N+1)
\eea respectively, where $\tb{I}$ denotes the identity matrix. Each
matrix $\tb{G}^{(n)}_{k}$ of dimension $\Delta_{n-1}\times\Delta_n$
is {\color{black}the compression of the training tensor
$\bm{\mc{X}}^{(k)}$.}
 The parameters $\Delta_{j}$ are called the bond dimensions or compression ranks of the MPS.  Using the common factors $\tb{B}^{(j)}_{i_{j}}$ and $\tb{C}^{(j)}_{i_{(j-1)}}$, we can extract the core {\color{black}matrices} for the test tensors
 $\bm{\mc{Y}}^{(\ell)}$ {\color{black} as follows.} We permute all $\bm{\mc{Y}}^{(\ell)}$, $\ell=1,\cdots,L$
 to ensure the compatibility between the training and test tensors. The compressed matrix  $\tb{Q}^{(n)}_{\ell}\in\mathbb{R}^{\Delta_{n-1}\times\Delta_n}$ of the test tensor $\bm{\mc{Y}}^{(\ell)}$ is then given by{\color{black}
\bea
\label{testcoreExtract}
\tb{Q}^{(n)}_{\ell}&=&\sum_{i_1,\ldots, i_{N}}(\tb{B}^{(1)}_{i_1})^{T}\cdots (\tb{B}^{(n-1)}_{i_{n-1}})^{T}y^{(\ell)}_{i_1\cdots \cdots i_{N}}\nonumber\\
&&(\tb{C}^{(n+1)}_{i_{n}})^{T}\cdots (\tb{C}^{(N+1)}_{i_{N}})^{T}.
\eea}
{\color{black} The dimension
\bea
N_f=\De_{n-1}\De_n
\label{NfMPS}
\eea
is the number of reduced features.}
\section{Tailored MPS for tensor compression}\label{MPSalgorithms}
The advantage of MPS for tensor compression is that the order $N$ of a tensor does not affect directly the feature
number $N_f$ in Eq. (\ref{NfMPS}), which is only determined strictly by the product
of the aforementioned bond dimensions $\De_{n-1}$ and $\De_n$.  {\color{black}In order to keep  $\De_{n-1}$ and $\De_n$
{\color{black}to a moderate size}, it is important to control the bond dimensions $\Delta_j$, and also to optimize
the positions of tensor modes as we address in this section.} {\color{black}In what follows, for
a matrix $\bm{X}$ we denote $\bm{X}(i,:)$ ($\bm{X}(:,j)$, resp.) as its $i$th row ($j$th column, resp.), while
for a third-order tensor $\bm{\mc{X}}$ we denote $\bm{\mc{X}}(:,\ell,:)$ as a matrix such that its
$(i_1,i_3)$th entry is {\color{black}$\bm{\mc{X}}(i_1,\ell,i_3)$}. For a $N$th-order tensor $\bm{\mc{X}}\in\mathbb{R}^{I_1\times\cdots\times I_N}$ we denote $\tb{X}_{[j]}\in\mathbb{R}^{(I_1I_2\cdots I_j)\times(I_{j+1}\cdots K \cdots I_{N})}$ as  its
\emph{mode-$(1,2,\ldots, j)$ matricization}. It is obvious that $\tb{X}_{[1]}=\tb{X}_{(1)}$.}

\subsection{Adaptive bond dimension control in MPS}\label{MPS2}
To decompose the training tensor $\bm{\mc{X}}$ into the MPS
according to Eq.~(\ref{mpsdef_mixed}), we apply two successive
sequences of SVDs to the tensor which include left-to-right sweep
for computing the left common factors
$\tb{B}^{(1)}_{i_1},\ldots,\tb{B}^{(n-1)}_{i_{n-1}}$, and
right-to-left sweep for computing the right common factors
$\tb{C}^{(n+1)}_{i_{n}},\ldots,\tb{C}^{(N+1)}_{i_{N}}$ and the core
matrix $\tb{G}^{(n)}_k$ in Eq.~(\ref{mpsdef_mixed}) as follows:

\textbullet~~\emph{Left-to-right sweep for left factor computation:}

The left-to-right sweep involves acquiring matrices $\tb{B}^{(j)}_{i_j}$ ($i_j=1,\ldots,I_j;\ j = 1,\ldots,n-1$) fulfilling orthogonality condition in Eq. (\ref{LeftCan1}).
Start by performing the mode-1 matricization of $\bm{\mc{X}}$ to obtain  \[
\tb{W}^{(1)}:=\bm{\mc{X}}_{[1]}=\bm{\mc{X}}_{(1)}\in\mb{R}^{I_1\times(I_2\cdots K\cdots I_{N})}.
\]
For
\begin{equation}\label{ra1}
\De_1\leq\text{rank}(\tb{X}_{[1]}),
\end{equation}
apply SVD to $\tb{W}^{(1)}$ to have the QR-approximation{\color{black}
\bea\label{ra2}
\tb{W}^{(1)}&\approx& \tb{U}^{(1)}\tb{V}^{(1)}\in\mathbb{R}^{I_1\times(I_2\cdots K\cdots I_{N})},
\eea
where $\tb{U}^{(1)}\in\mathbb{R}^{I_1\times \Delta_1}$ is orthogonal:
\begin{equation}\label{ra2a}
(\tb{U}^{(1)})^T\tb{U}^{(1)}=\tb{I},
\end{equation}
and $\tb{V}^{(1)}\in\mathbb{R}^{\Delta_1\times(I_2\cdots K\cdots I_{N})}$.}
Define the {\color{black}the most left} common factors by
\begin{equation}\label{ra3}
\tb{B}^{(1)}_{i_1} = \tb{U}^{(1)}(i_1,:)\in\mb{R}^{1\times\De_1}, i_1=1,\cdots,I_1
\end{equation}
which satisfy  the left-canonical constraint in Eq. (\ref{LeftCan1}) due to  (\ref{ra2a}). \\
{\color{black}Next, reshape the matrix} $\tb{V}^{(1)}\in\mb{R}^{\De_1\times(I_2\cdots K\cdots I_{N})}$
to $\tb{W}^{(2)}\in\mb{R}^{(\De_1I_2)\times(I_3\cdots K\cdots I_{N})}$.
For
\begin{equation}\label{ra5}
\De_2\leq\text{rank}(\tb{W}^{(2)})\leq \text{rank}(\tb{X}_{[2]}),
\end{equation}
apply SVD to $\tb{W}^{(2)}$ for the QR-approximation{\color{black}
\begin{equation}\label{ra4}
\tb{W}^{(2)}\approx \tb{U}^{(2)}\tb{V}^{(2)}\in\mathbb{R}^{(\De_1I_2)\times(I_3\cdots K\cdots I_{N})},
\end{equation}
where $\tb{U}^{(2)}\in\mathbb{R}^{(\De_1I_2)\times \Delta_2}$ is orthogonal such that
\begin{equation}\label{ra4a}
(\tb{U}^{(2)})^T\tb{U}^{(2)}=\tb{I},
\end{equation}
and $\tb{V}^{(2)}\in\mathbb{R}^{\De_2\times(I_3\cdots K\cdots I_{N})}$.}
Reshape the matrix $\tb{U}^{(2)}\in\mb{R}^{(\De_1I_2)\times\De_2}$
into a third-order tensor $\bm{\mc{U}}\in\mb{R}^{\De_1\times I_2\times\De_2}$ to
define the next common factors
\begin{equation}\label{ra6}
\tb{B}^{(2)}_{i_2} = \bm{\mc{U}}(:,i_2,:)\in
\mathbb{R}^{\Delta_1\times \Delta_2}, i_2=1,\cdots,I_2,
\end{equation}
which satisfy the left-canonical constraint due to  (\ref{ra4a}). \\
Applying the same procedure for determining $\tb{B}^{(3)}_{i_3}$ by reshaping the matrix
$\tb{V}^{(2)}\in\mb{R}^{\De_2\times(I_3\cdots K\cdots I_{N})}$
to
\[
\tb{W}^{(3)}\in\mb{R}^{(\De_2I_3)\times(I_4\cdots K\cdots I_{N})},
\]
{\color{black} performing the SVD, and so on}. This procedure is iterated till obtaining the last QR-approximation{\color{black}
\begin{equation}\label{ra7}
\begin{array}{lll}
\tb{W}^{(n-1)}&\approx&
\tb{U}^{(n-1)}\tb{V}^{(n-1)}\in\mathbb{R}^{(\Delta_{n-2}I_{n-1})\times
(KI_{n}\cdots I_N)},\\
&&\tb{U}^{(n-1)}\in\mathbb{R}^{(\Delta_{n-2}I_{n-1})\times\Delta_{n-1}},\\
&&\tb{V}^{(n-1)}\in\mathbb{R}^{\Delta_{n-1} \times (KI_{n}\cdots
I_N)},
\end{array}
\end{equation}}
with $\tb{U}^{(n-1)}$ orthogonal:
\begin{equation}\label{ra7a}
\tb{U}^{(n-1)}(\tb{U}^{(n-1)})^T=\tb{I}
\end{equation}
and reshaping $\tb{U}^{(n-1)}\in\mathbb{R}^{(\Delta_{n-2}I_{n-1})\times\Delta_{n-1}}$
into a third-order tensor $\bm{\mc{U}}\in\mb{R}^{\De_{n-2}\times I_{n-1}\times\De_{n-1}}$
to define the last left common factors
\begin{equation}\label{ra8}
\tb{B}^{(n-1)}_{i_{n-1}} = \bm{\mc{U}}(:,i_{n-1},:)\in \mathbb{R}^{\Delta_{n-2}\times \Delta_{n-1}}, i_{n-1}=1,\cdots, I_{n-1},
\end{equation}
which satisfy the left-canonical constraint due to  (\ref{ra7a}).

In a nutshell, after completing the left-to-right sweep, the elements of
tensor $\bm{\mc{X}}$ are approximated by \bea x^{(k)}_{i_1\cdots
i_{n-1}i_n\cdots i_{N+1}}\approx\tb{B}^{(1)}_{i_1}\cdots
\tb{B}^{(n-1)}_{i_{n-1}} \tb{V}^{(n-1)} (:,ki_n\cdots i_{N}).
\label{l7} \eea The matrix
$\tb{V}^{(n-1)}\in\mathbb{R}^{\Delta_{n-1}\times (KI_n\cdots I_N)}$
is reshaped to $\tb{W}^{(N)}\in\mb{R}^{(\De_{n-1}K\cdots
I_{N-1})\times I_{N}}$ for the next right-to-left sweeping process.

\textbullet~~\emph{Right-to-left sweep for right factor computation:}

Similar to left-to-right sweep, we perform a sequence of SVDs starting from the right to the left of the MPS to get the  matrices $\tb{C}^{(j)}_{i_{j-1}}$ ($i_{j-1}=1,\ldots,I_{j-1}$; $j = N+1,\ldots,n+1$) fulfilling the right-canonical condition in Eq. (\ref{RightCan1}). To start, we apply the SVD to the
matrix $\tb{W}^{(N)}\in\mb{R}^{(\De_{n-1}K\cdots I_{N-1})\times I_{N}}$ obtained previously in the left-to-right sweep to have the RQ-approximation{\color{black}
\begin{equation}\label{ri1}
\tb{W}^{(N)}\approx \tb{U}^{(N)}\tb{V}^{(N)},
\end{equation}
where $\tb{U}^{(N)}\in\mathbb{R}^{(\De_{n-1}K\cdots I_{N-1})\times\Delta_N}$ and $\tb{V}^{(N)}\in\mathbb{R}^{\Delta_N\times I_N}$
is orthogonal:
\begin{equation}\label{ri1a}
\tb{V}^{(N)}(\tb{V}^{(N)})^T=\tb{I}
\end{equation}
}
for
\begin{equation}\label{ri2}
\De_N \leq \text{rank}(\tb{W}^{(N)})\leq {\color{black}\text{rank}(\bm{\mc{X}}_{[N-1]}).}
\end{equation}
Define {\color{black}the most right common factors}
\[\tb{C}^{(N+1)}_{i_{N}} = \tb{V}^{(N)}(:,i_{N})\in\mb{R}^{\De_N\times 1}, i_N=1,\cdots, I_N,
\]
which {\color{black}satisfy} the right-canonical constraint (\ref{RightCan1}) due to (\ref{ri1a}).\\
Next, reshape $\tb{U}^{(N)}\in\mathbb{R}^{(\De_{n-1}K\cdots I_{N-1})\times\Delta_N}$ into
$\tb{W}^{(N-1)}\in\mb{R}^{(\De_{n-1}K\cdots I_{N-2})\times (I_{N-1}\De_N)}$ and apply the SVD to  have
the RQ-approximation{\color{black}
\begin{equation}\label{ri3}
\tb{W}^{(N-1)}\approx \tb{U}^{(N-1)}\tb{V}^{(N-1)},
\end{equation}
where $\tb{U}^{(N-1)}\in\mathbb{R}^{(\De_{n-1}K\cdots I_{N-2})\times\De_{N-1}}$ and $\tb{V}^{(N-1)}\in\mathbb{R}^{\Delta_{N-1}\times(I_{N-1}\Delta_N)}$ is
orthogonal:
\begin{equation}\label{ri3a}
\tb{V}^{(N-1)}(\tb{V}^{(N-1)})^T=\tb{I}
\end{equation}}
for
\begin{equation}\label{ri3}
\De_{N-1} \leq \text{rank}(\tb{W}^{(N-1)})
{\color{black}\leq\text{rank}(\bm{\mc{X}}_{[N-2]})}.
\end{equation}
Reshape the matrix $\tb{V}^{(N-1)}\in\mb{R}^{\De_{N-1}\times(I_{N-1}\De_N)}$  into a third-order tensor
$\bm{\mc{V}}\in\mb{R}^{\De_{N-1}\times I_{N-1}\times\De_N}$ to
define the next common factor
\begin{equation}\label{ri4}
\tb{C}^{(N)}_{i_{N-1}} =
\bm{\mc{V}}(:,i_{N-1},:)\in\mathbb{R}^{\De_{N-1}\times\De_N}
\end{equation}
which satisfy Eq. (\ref{RightCan1}) due to (\ref{ri3a}).\\
This procedure is iterated till  obtaining the last RQ-approximation
\begin{equation}\label{ri5}
\begin{array}{lll}
\tb{W}^{(n)}&\approx&
\tb{U}^{(n)}\tb{V}^{(n)}\in\mathbb{R}^{(\Delta_{n-1}K)\times
(I_{n}\Delta_{n+1})},\\
&&\tb{U}^{(n)}\in\mathbb{R}^{(\Delta_{n-1}K)\times\Delta_{n}},
\tb{V}^{(n)}\in\mathbb{R}^{\Delta_{n} \times (I_{n}\Delta_{n+1})},
\end{array}
\end{equation}
with $\tb{V}^{(n)}$ orthogonal:
\begin{equation}\label{ri5a}
\tb{V}^{(n)}(\tb{V}^{(n)})^T=\tb{I}
\end{equation}
for
\begin{equation}\label{ri6}
\De_{n} \leq \text{rank}(\tb{W}^{(n)})
{\color{black}\leq\text{rank}(\bm{\mc{X}}_{[n-1]})}.
\end{equation}
Reshape
$\tb{V}^{(n)}\in\mathbb{R}^{(\Delta_{n})\times(I_n\Delta_{n+1})}$
into a third-order tensor $\bm{\mc{V}}\in\mb{R}^{\De_{n}\times
I_{n}\times\De_{n+1}}$ to define the last right common factors
\begin{equation}\label{ri7}
\tb{C}^{(n+1)}_{i_{n}} = \bm{\mc{V}}(:,i_{n},:)\in
\mathbb{R}^{\Delta_{n-1}\times \Delta_{n}}, i_{n}=1,\cdots, I_{n},
\end{equation}
which satisfy  (\ref{RightCan1}) due to  (\ref{ri5a}).

By reshaping
$\tb{U}^{(n)}\in\mathbb{R}^{(\Delta_{n-1}K)\times\Delta_{n}}$ into a
third-order tensor $\bm{\mc{G}}\in\mathbb{R}^{\Delta_{n-1}\times
K\times\Delta_{n}}$ to define $\tb{G}^{(n)}_{k}=\bm{\mc{G}}(:,k,:)$,
$k=1,\cdots.K$,
 {\color{black}we arrive at  Eq.~(\ref{mpsdef_mixed}).}

Note that the MPS decomposition described by Eq. (\ref{mpsdef_mixed}) can be performed exactly or approximately depending on the bond dimensions $\Delta_{j}$ $(j=1,\ldots, N)$. The bond dimension truncation  is of crucial importance to control
the final feature number $N_f=\Delta_{n-1}\Delta_n$. To this end, we rely on thresholding the singular values of $\tb{W}^{(j)}$.
With a threshold $\epsilon$ being defined in advance, we control $\De_j$ such that $\De_j$ largest
singular values $s_1\geq s_2\geq ...\geq s_{\Delta_j}$ satisfy
\bea
\frac{\sum_{i=1}^{\De_j}s_i}{\sum_{i=1}^{r_j}s_j}\geq\epsilon,
\label{BondThreshold}
\eea
for $r_j=\text{rank}(\tb{W}^{(j)})$. {\color{black}The information loss from the von Neumann entropy (\ref{tdloss2}) of $\tb{W}^{(j)}$ by this truncation is given by (\ref{tdloss1}).  The entropy of each $\tb{W}^{(j)}$ provides the correlation degree between two sets of modes $1,\cdots,j$ and $j+1,\cdots,N$ \cite{BENGUA2016}. Therefore, the $N$ entropies $\tb{W}^{(j)}$, $j=1,\cdots,N$ provide the mean of the tensor's global correlation. Furthermore, rank $r_j$ of each $\tb{W}^{(j)}$ is upper bounded by
\begin{equation}\label{TTr}
\min\ \{I_1\cdots I_j, I_{j+1}\cdots I_N\}
\end{equation}
making the truncation (\ref{BondThreshold}) highly favorable in term of compression loss to matrices of higher rank due to
balanced row and column numbers.}\\
A detailed outline of our MPS approach to tensor feature extraction is presented in Algorithm 1.
\begin{table}[!thb]
    \centering
    \caption*{Algorithm I: MPS for tensor feature extraction}
    \label{Train_MPS}
    \begin{tabular}{*2l} 
        \hline
        \tb{Input:}& $\bm{\mc{X}}\in\mathbb{R}^{I_1\times \cdots \times I_{n-1}\times K\cdots\times I_{N}}$,\\
        &$\epsilon$: SVD threshold\\
        \tb{Output:}& $\tb{G}_k^{(n)}\in\mathbb{R}^{\De_{n-1}\times\De_{n}}$, $k=1,\cdots,K$\\
                   ~& $\tb{B}^{(j)}_{i_j}$ ($i_j=1,\ldots,I_j, j = 1,\ldots,n-1$)\\
                   ~& $\tb{C}^{(j)}_{i_{(j-1)}}$ ($i_{(j-1)}=1,\ldots,I_{(j-1)}, j = n+1,\ldots,N+1$)\\
        \hline
        \multicolumn{2}{l}{1:~~Set $\tb{W}^{(1)}=\tb{X}_{(1)}$~~~~~~~~~~~~~$\%$ Mode-1 matricization of $\bm{\mc{X}}$}\\
        \multicolumn{2}{l}{2:~~\tb{for} $j=1$ \tb{to} $n-1$ ~~~~$\%$ Left-to-right sweep}\\
        \multicolumn{2}{l}{3:~~~~~~$\tb{W}^{(j)}= \tb{U}\tb{S}\tb{V}$~~~~~~~~~~$\%$ SVD of $\tb{W}^{(j)}$}\\
        \multicolumn{2}{l}{4:~~~~~~$\tb{W}^{j}\approx \tb{U}^{(j)}\tb{W}^{(j+1)}$~~~~~~~~~~$\%$ Thresholding $\tb{S}$ for QR-approximation}\\
        \multicolumn{2}{l}{5:~~~~~Reshape $\tb{U}^{(j)}$ to $\bm{\mc{U}}$}\\
        \multicolumn{2}{l}{6:~~~~~~$\tb{B}^{(j)}_{i_j} = \bm{\mc{U}}(:,i_j,:)$~~~~~~~~~~~~$\%$ Set common factors}\\
        \multicolumn{2}{l}{7:~~\tb{end}}\\
        \multicolumn{2}{l}{8:~~Reshape $\tb{V}^{(n-1)}$ to
        $\tb{W}^{N}\in\mb{R}^{(\De_{n-1}K\cdots I_{N})\times I_{N}}$}\\
        \multicolumn{2}{l}{9:~~\tb{for} $j=N$ \tb{down to} $n$ ~~$\%$ right-to-left sweep}\\
        \multicolumn{2}{l}{10:~~~~~$\tb{W}^{(j)}= \tb{U}\tb{S}\tb{V}$~~~~~~~~~$\%$ SVD of $\tb{W}^{(j)}$}\\
        \multicolumn{2}{l}{11:~~~~~$\tb{W}^{(j)}\approx \tb{W}^{(j-1)}\tb{V}^{(j)}$~~~~~~~~~$\%$ Thresholding $\tb{S}$
        for RQ-approximation}\\
        \multicolumn{2}{l}{13:~~~~~Reshape $\tb{V}^{(j)}$ to $\bm{\mc{V}}$}\\
        \multicolumn{2}{l}{14:~~~~~$\tb{C}^{(j+1)}_{i_{j-1}} = \bm{\mc{V}}(:,i_{j-1},:)$~~~~~~~~~~$\%$ Set common factors}\\
        \multicolumn{2}{l}{15:~\tb{end}}\\
        \multicolumn{2}{l}{16:~ Reshape $\tb{U}^{(n)}$ into  $\bm{\mc{G}}\in\mathbb{R}^{\Delta_{n-1}\times
K\times\Delta_{n}}$} \\
        \multicolumn{2}{l}{17:~Set $\tb{G}^{(n)}_{k}=\bm{\mc{G}}(:,k,:)$~~~~~~~~~~~~~~  $\%$ Training core matrix}\\
        \hline
        \multicolumn{2}{l}{Texts after symbol "$\%$" are comments.}
    \end{tabular}
\end{table}
\subsection{Tensor mode pre-permutation and pre-positioning mode $K$ for MPS}
One can see from (\ref{TTr}) that the efficiency of controlling the bond dimension $\Delta_j$ is dependent on its upper bound
(\ref{TTr}). Particularly, the efficiency of controlling the {\color{black}bond} dimensions $\Delta_{n-1}$ and $\Delta_n$ that define the
feature number (\ref{NfMPS}) is dependent on
\begin{equation}\label{TTr1}
\min\ \{I_1\cdots I_{n-1}, I_{n}\cdots I_N\}
\end{equation}
Therefore, it is important to pre-permute the tensors modes such that the ratio
\begin{equation}\label{TTr2}
\frac{\min\{\prod_{i=1}^{n-1}I_i,\prod_{i=n}^{N}I_i\}}{\max\{\prod_{i=1}^{n-1}I_i,\prod_{i=n}^{N}I_i\}}
\end{equation}
is near to $1$ as possible, while $\{I_1,\cdots,I_{n-1}\}$ is in decreasing order
\begin{equation}\label{TTr3}
I_1\geq\cdots\geq I_{n-1}
\end{equation}
and $\{I_n,\cdots,I_N\}$ in increasing order
\begin{equation}\label{TTr3}
I_n\leq\cdots\leq I_N
\end{equation}
to improve the ratio
\begin{equation}\label{TTr3}
\frac{\min\{\prod_{i=1}^{j}I_j,\prod_{i=j+1}^{N}I_i\}}{\max\{\prod_{i=1}^{j}I_j,\prod_{i=j+1}^{N}I_i\}}
\end{equation}
for balancing $\tb{W}^{(j)}$.  \\
The mode $K$ is then pre-positioned in $n$-th mode as in (\ref{pos}).
\subsection{Complexity analysis}\label{Complexity}
In the following complexity  analysis it is assumed $I_n = I$ $\forall n$ for simplicity.
The dominant computational complexity of MPS is $\mc{O}(KI^{(N+1)})$ due to the first SVD of the matrix obtained from the mode-1 matricization of $\bm{\mc{X}}$. On the other hand, the computational complexity of HOOI requires several iterations of an ALS method to obtain convergence. In addition, it usually employs the HOSVD to initialize the tensors which involves the cost of order $\mc{O}(NKI^{N+1})$, and thus very expensive with large $N$ compared to MPS.

MPCA is computationally upper bounded by $\mc{O}(NKI^{N+1})$,
however, unlike HOOI, MPCA doesn't require the formation of the
$(N+1)$th order core tensor at every iteration and convergence can
usually happen in one iteration \cite{LU08}.\footnote{This does not
mean that MPCA is computationally efficient but in contrast this
means that alternating iterations of MPCA prematurely terminate,
yielding a solution that is far from the optimal one of a NP-hard problem.}

The computational complexity of R-UMLDA is approximately $\mc{O}(K\sum_{n=2}^N I^n + (C+K)I^2 + (p-1)[IK+2I^2+(p-1)^2+(2I(p-1)]+4I^3)$, where $C$ is the number of classes, $p$ is the number of projections, which determines the core vector size \cite{4711317}. Therefore, R-UMLDA would perform poorly for many samples and classes.

\subsection{MPS-based tensor object classification}\label{mpsbtoc}
This subsection presents two methods for tensor objection classification based on Algorithm 1. For each method, an explanation of how to reduce the dimensionality of tensors to core matrices, and subsequently to feature vectors for application to linear classifiers is given.
\subsubsection{Principal component analysis via tensor-train (TTPCA)} The TTPCA algorithm is an approach where Algorithm 1 is applied directly on the training set, with no preprocessing such as data centering. Specifically, given a set of $N$th-order tensor samples $\mathcal{X}^{(k)}\in\mathbb{R}^{I_1\times I_2\times\cdots\times I_N}$, then the core matrices are obtained as
\bea
\tb{G}^{(n)}_{k}\in\mathbb{R}^{\De_{n-1}\times\De_n}.
\eea
Vectorizing each $k$ sample results in
\bea
\tb{g}^{(n)}_{k}\in\mathbb{R}^{\De_{n-1}\De_n}.\label{veccore}
\eea
Using (\ref{BondThreshold}), $\De_{n-1}\De_n$ features of $k$ is significantly less in comparison to $N_f=\prod_{n=1}^N I_n$ of $\mathcal{X}^{(k)}$, which allows for PCA to be easily applied, followed by a linear classifier.

\subsubsection{MPS}The second algorithm is simply called MPS, where in this case we first perform data centering on the set of training samples $\{\bm{\mc{X}}^{(k)} \}$, then apply Algorithm 1 to obtain the core matrices
\bea
\tb{G}^{(n)}_{k}\in\mathbb{R}^{\De_{n-1}\times\De_n}.
\eea
Vectorizing the $K$ samples results in (\ref{veccore}), and subsequent linear classifiers such as LDA or nearest neighbors can be utilized. In this method, MPS can be considered a multidimensional analogue to PCA because the tensor samples have been data centered and are projected to a new orthogonal space using Algorithm 1, resulting in the core matrices.

\section{Experimental results}\label{Ressec}
{\color{black}In this section, we conduct experiments on the proposed TTPCA and MPS algorithms for tensor object classification. An extensive comparison is conducted based on CSR and training time with tensor-based methods MPCA, HOOI, and R-UMLDA.}

Four datasets are utilized for the experiment. The Columbia Object Image Libraries (COIL-100) \cite{NEN96b,PON98}, Extended Yale Face Database B (EYFB) \cite{GEO01}, BCI Jiaotong dataset (BCI) \cite{BCM13}, and the University of South Florida HumanID ``gait challenge`` dataset (GAIT) version 1.7 \cite{1374864} . All simulations are conducted in a Matlab environment.

\subsection{Parameter selection}\label{paramdisc}
TTPCA, MPA and HOOI rely on the threshold $\epsilon$ defined in (\ref{BondThreshold}) to reduce the dimensionality of a tensor, while keeping its most relevant features. To demonstrate how the classification success rate (CSR) varies, we utilize different  $\epsilon$ for each dataset. It is trivial to see that a larger $\epsilon$ would result in a longer training time due to its computational complexity, which was discussed in subsection \ref{Complexity}. Furthermore, TTPCA utilizes PCA, and a range of principal components $p$ is used for the experiments. HOOI is implemented with a maximum of 10 ALS iterations.
MPCA relies on fixing an initial quality factor $Q$, which is determined through numerical simulations, and a specified number of elementary multilinear projections (EMP), we denote as $m_p$, must be initialized prior to using the R-UMLDA algorithm. A range of EMP's is determined through numerical simulations and the regularization parameter is fixed to $\gamma=10^{-6}$ .

\begin{table*}[htpb!]
    \caption{COIL-100 classification results. The best CSR corresponding to different H/O ratios obtained by MPS and HOOI.}
    \label{tableCOIL100}
    \centering 
    \begin{tabular}{l l l l |l l l}
Algorithm & CSR & $N_f$& $\epsilon$ & CSR & $N_f$& $\epsilon$\\
        \hline
        \underline{$r=50\%$}&&&&\underline{$r=80\%$}&&\\
        HOOI & $98.87\pm 0.19$ & $198$ & $0.80$&$94.13\pm 0.42$&$112$&$0.75$\\
        MPS & $\bf{99.19\pm 0.19}$ & $120$ & $0.80$&$\bf{95.37\pm 0.31}$&$18$&$0.65$\\   \underline{$r=90\%$}&&&&\underline{$r=95\%$}&&\\
    HOOI & $87.22\pm 0.56$ & $112$&$0.75$&$77.76\pm 0.90$&$112$&$0.75$\\
        MPS& $\bf{89.38\pm 0.40}$ & $59\pm 5$&$0.75$&${\bf 83.17\pm 1.07}$&$18$&$0.65$\\
        \hline
    \end{tabular}
\end{table*}

\subsection{Tensor object classification}
\subsubsection{COIL-100}
\begin{figure}[htpb!]
    \centering
    \includegraphics[width =\columnwidth]{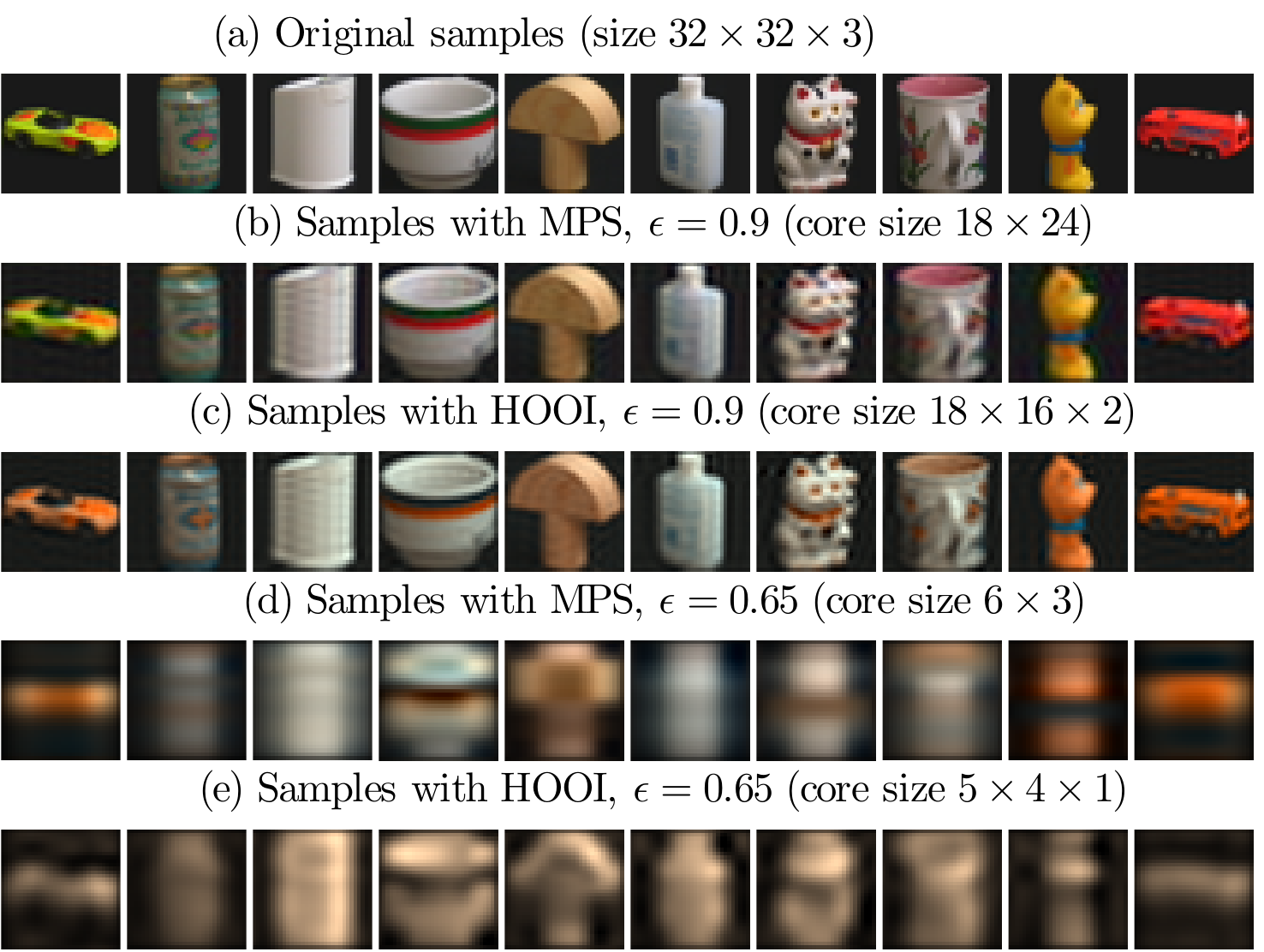}\\
    \caption{Modification of ten objects in the training set of COIL-100 are shown after applying MPS and HOOI corresponding to $\epsilon=0.9$ and $0.65$ to compress  tensor objects.}
    \label{fig1}
\end{figure}

{\color{black}For this dataset we strictly compare MPS and the HOSVD-based algorithm HOOI to analyse how adjusting $\epsilon$ affects the approximation of the original tensors, as well as the reliability of the extracted features for classification.} The COIL-100 dataset has 7200 color images of 100 objects (72 images per object) with different
reflectance and complex geometric characteristics.
Each image is initially a 3rd-order tensor of dimension $128\times128\times3$ and then is downsampled to the one of dimension $32\times32\times3$. The dataset is divided into training and test sets randomly consisting of $K$ and $L$ images, respectively according to a certain holdout (H/O) ratio $r$, i.e. $r=\frac{L}{K}$. Hence, the training and test sets are represented by four-order tensors of dimensions $32\times32\times3\times K$ and $32\times32\times3 \times L$, respectively.
{\color{black}In Fig.~\ref{fig1} we show how a few objects of the training  set ($r=0.5$ is chosen) change after {\color{black}compression} by
MPS and HOOI  with two different values of threshold, $\epsilon=0.9,0.65$.} We can see that with $\epsilon=0.9$, the images are not modified significantly due to the fact that many features are preserved. However, in the case that $\epsilon=0.65$, the images are blurred. That is because fewer features are kept. However, we can observe that the shapes of objects are still preserved. Especially, in most cases MPS seems to preserve the color of the images better than HOOI. This is because the bond dimension corresponding to the color mode $I_3=3$ has a small value, e.g. $\De_3=1$ for $\epsilon=0.65$ in HOOI. This problem arises due to the  the unbalanced matricization of the tensor corresponding to the color mode. Specifically, if we take a mode-3 matricization of tensor $\bm{\mc{X}}\in\mb{R}^{32\times32\times3\times K}$, the resulting matrix of size $3\times(1024K)$ is extremely unbalanced. Therefore, when taking SVD with some small threshold $\epsilon$, the information corresponding to this color mode may be lost due to dimension reduction. On the contrary, we can efficiently avoid this problem in MPS by permuting the tensor such that $\bm{\mc{X}}\in\mb{R}^{32\times K\times3\times32}$ before applying the tensor decomposition.

\begin{figure}[htpb!]
    \centering
    \includegraphics[width =\columnwidth]{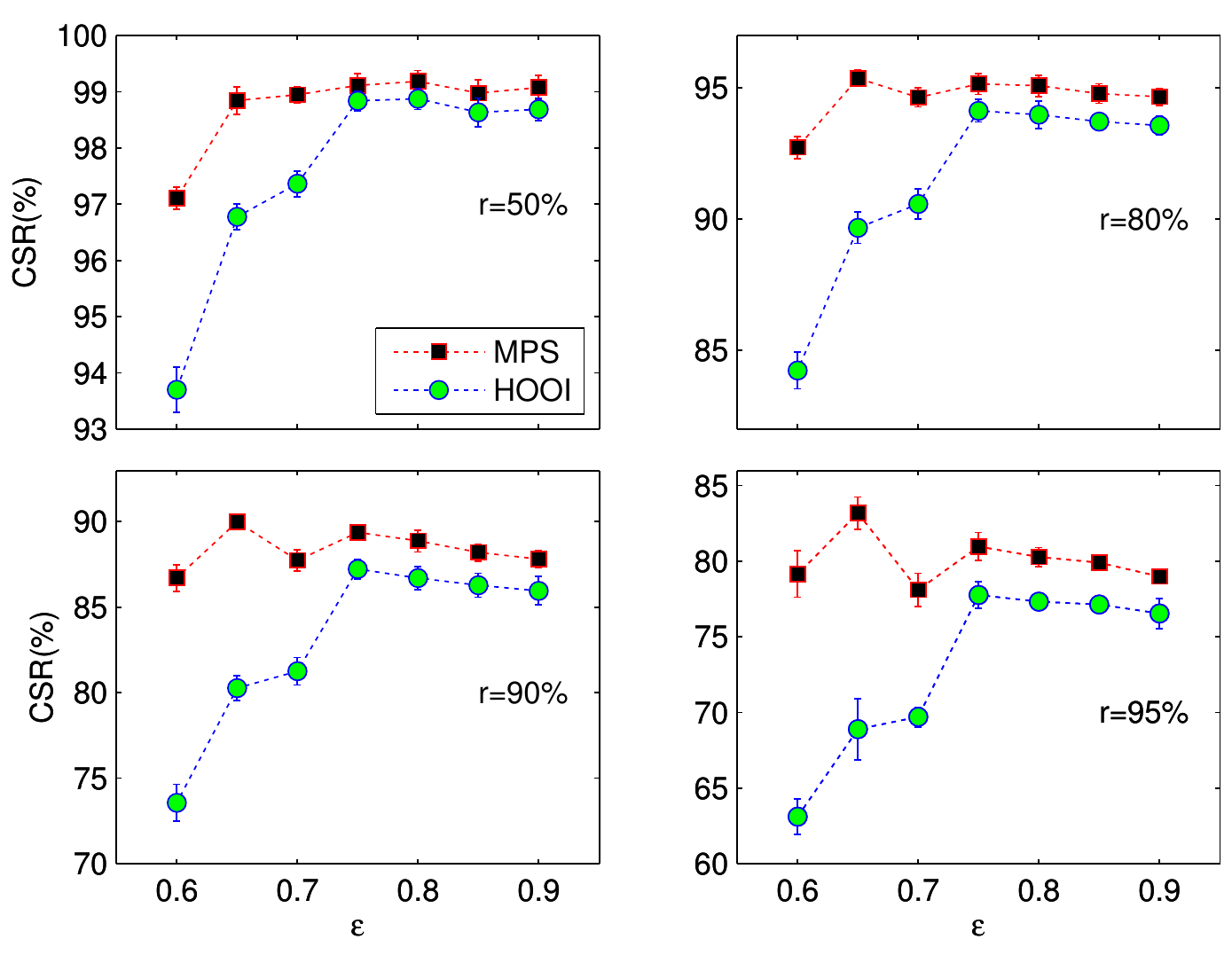}\\
    \caption{Error bar plots of CSR versus thresholding rate $\epsilon$ for different H/O ratios.}
    \label{fig3}
\end{figure}

{\color{black}K nearest neighbors with K=1 (KNN-1) is used for classification. For each H/O ratio, the CSR} is averaged over 10 iterations of randomly splitting the dataset into training and test sets. Comparison of performance between MPS and HOOI is shown in Fig.~\ref{fig3} for four different H/O ratios, i.e. $r=(50\%,80\%,90\%,95\%)$. In each plot, we show the CSR with respect to threshold $\epsilon$. We can see that MPS performs quite well when compared to HOOI. Especially, with small $\epsilon$, MPS performs much better than HOOI. Besides, we also show the best CSR corresponding to each H/O ratio obtained by different methods in Table.~\ref{tableCOIL100}. It can be seen that MPS always gives better results than HOOI even in the case of small value of $\epsilon$ and number of features $N_f$
defined by (\ref{NfTD}) and (\ref{NfMPS}) for HOOI and MPS, respectively.

\subsubsection{Extended Yale Face Database B}
The EYFB dataset contains 16128 grayscale images with 28 human subjects, under 9 poses, where for each pose there is 64 illumination conditions. Similar to \cite{LI14}, to improve computational time each image was cropped to keep only the center area containing the face, then resized to 73 x 55. The training and test datasets are not selected randomly but partitioned according to poses. More precisely, the training and test datasets are selected to contain poses 0, 2, 4, 6 and 8 and 1, 3, 5, and 7, respectively. For a single subject the training tensor has size $5\times73\times55\times64$ and $4\times73\times55\times64$ is the size of the test tensor. Hence for all 28 subjects we have fourth-order tensors of sizes $140\times73\times55\times64$ and $112\times73\times55\times64$ for the training and test datasets, respectively.
\begin{table*}[!htbp]
    \caption{EYFB classification results}
    \label{tablecEYFB}
    \centering 
    \begin{tabular}{l l l l l}
        \multirow{1}{*}{Algorithm} & CSR ($\epsilon = 0.9$)&  CSR ($\epsilon = 0.85$)&
        CSR ($\epsilon = 0.80$)
        & CSR ($\epsilon = 0.75$) \\
        \hline
        {\bf KNN-1}&&&&\\
        HOOI & $90.71\pm 1.49$ & $90.89\pm 1.60$ & $91.61\pm 1.26$ & $88.57\pm 0.80$\\
        MPS & ${\bf 94.29\pm 0.49}$ & ${\bf94.29\pm 0.49}$ & ${\bf 94.29\pm 0.49}$ & ${\bf 94.29\pm 0.49}$\\
        TTPCA & $86.05\pm 0.44$ & $86.01\pm 0.86$ & $87.33\pm 0.46$ & $86.99\pm 0.53$\\
        MPCA & $90.89\pm 1.32$ \\
        R-UMLDA & $71.34\pm 2.86$\\
        \hline
        {\bf LDA}&&&&\\
        HOOI & $96.07\pm 0.80$ & $95.89\pm 0.49$ & $96.07\pm 0.49$ & $96.07\pm 0.49$\\
        MPS & ${\bf 97.32\pm 0.89}$ & ${\bf97.32\pm 0.89}$ & ${\bf 97.32\pm 0.89}$ & ${\bf 97.32\pm 0.89}$\\
        TTPCA & $95.15\pm 0.45$ & $95.15\pm 0.45$ & $95.15\pm 0.45$ & $94.86\pm 0.74$\\
        MPCA & $90.00\pm 2.92$\\
        R-UMLDA & $73.38\pm 1.78$\\
        \hline
    \end{tabular}
\end{table*}

\begin{table*}[!ht]
    \caption{BCI Jiaotong classification results}
    \label{tablecBCI}
    \centering 
    \begin{tabular}{l l l l l}
        \multirow{1}{*}{Algorithm} & CSR ($\epsilon = 0.9$)&  CSR ($\epsilon = 0.85$)&
        CSR ($\epsilon = 0.80$)
        & CSR ($\epsilon = 0.75$) \\
        \hline
        {\bf Subject 1}& & & &\\
        HOOI & $84.39\pm 1.12$ & $83.37\pm 0.99$ & $82.04\pm 1.05$ & $84.80\pm 2.21$\\
        MPS & ${\bf 87.24\pm 1.20}$ & ${\bf 87.55\pm 1.48}$ & ${\bf 87.24\pm 1.39}$ & ${\bf 87.65\pm 1.58}$\\
        TTPCA & $ 78.57\pm 3.95$ & $ 78.43\pm 3.73$ & ${79.43\pm 4.12}$ & ${ 79.14\pm 2.78}$\\
        MPCA  & $82.14\pm 3.50$ \\
        R-UMLDA & $63.18\pm 0.37$ \\
        CSP & $80.14\pm 3.73$ \\
        \hline
        {\bf Subject 2}& & & &\\
        HOOI & $83.16\pm 1.74$ & $82.35\pm 1.92$ & $82.55\pm 1.93$ & $79.39\pm 1.62$\\
        MPS & ${\bf 90.10\pm 1.12}$ & ${\bf 90.10\pm 1.12}$ & ${\bf 90.00\pm 1.09}$ & ${\bf 91.02\pm 0.70}$\\
        TTPCA & ${ 80.57\pm 0.93}$ & ${ 81.14\pm 1.86}$ & ${ 81.29\pm 1.78}$ & ${ 80\pm 2.20}$\\
        MPCA & $81.29\pm 0.78$ \\
        R-UMLDA & $70.06\pm 0.39$\\
        CSP & $81.71\pm 8.96$ \\
        \hline
        {\bf Subject 3}& & & &\\
        HOOI & $60.92\pm 1.83$ & $61.84\pm 1.97$ & $61.12\pm 1.84$ & $60.51\pm 1.47$\\
        MPS & ${ 61.12\pm 1.36}$ & ${ 61.22\pm 1.53}$ & ${ 61.12\pm 1.54}$ & ${60.71\pm 1.54}$\\
        TTPCA & ${ 67.43\pm 2.56}$ & ${ 68.29\pm 2.56}$ & ${ 67.71\pm 2.28}$ & ${ 66.43\pm 2.02}$\\
        MPCA & $56.14\pm 2.40$\\
        R-UMLDA & $57.86\pm 0.00$\\
        CSP & ${\bf77.14\pm 2.26}$ \\
        \hline
        {\bf Subject 4}& & & &\\
        HOOI & $48.27\pm 1.54$ & $47.55\pm 1.36$ & $49.98\pm 1.29$ & $47.96\pm 1.27$\\
        MPS & ${ 52.35\pm 2.82}$ & ${ 52.55\pm 3.40}$ & ${ 52.55\pm 3.69}$ & ${ 51.84\pm 3.11}$\\
        TTPCA & ${ 50.29\pm 2.97}$ & ${ 49.71\pm 3.77}$ & ${ 49.14\pm 3.48}$ & ${ 52.00\pm 3.48}$\\
        MPCA & $51.00\pm 3.96$\\
        R-UMLDA & $46.36\pm 0.93$\\
        CSP & ${\bf 59.86\pm 1.98}$ \\
        \hline
        {\bf Subject 5}& & & &\\
        HOOI & ${\bf60.31\pm 1.08}$ & ${\bf60.82\pm 0.96}$ & ${\bf59.90\pm 2.20}$ & ${\bf60.41\pm 1.36}$\\
        MPS & ${59.39\pm 2.08}$ & ${ 59.18\pm 2.20}$ & ${ 58.57\pm 1.60}$ & ${ 59.29\pm 1.17}$\\
        TTPCA & ${ 53.43\pm 2.79}$ & ${ 54.29\pm 3.19}$ & ${ 53.86\pm 3.83}$ & ${ 54.86\pm 2.49}$\\
        MPCA & $50.43\pm 1.48$\\
        R-UMLDA & $55.00\pm 0.55$\\
        CSP & $59.14\pm 2.11$ \\
        \hline
    \end{tabular}
\end{table*}

\begin{table*}[!ht]
    \caption{GAIT classification results}
    \label{tablecGAIT}
    \centering 
    \begin{tabular}{l l l l l}
        \multirow{1}{*}{Algorithm} & CSR ($\epsilon = 0.9$)&  CSR ($\epsilon = 0.85$)&
        CSR ($\epsilon = 0.80$)
        & CSR ($\epsilon = 0.75$) \\
        \hline
        {\bf Probe A}& & & &\\
        HOOI & $63.71\pm 3.36$ & $63.90\pm 3.40$ & $64.16\pm 3.39$ & $64.33\pm 3.20$\\
        MPS & ${ 70.03\pm 0.42}$ & ${ 70.03\pm 0.38}$ & ${ 70.01\pm 0.36}$ & ${ 69.99\pm 0.38}$\\
        TTPCA & ${\bf 75.31\pm 0.29 }$ & ${\bf76.03\pm 0.38}$ & ${\bf76.38\pm 0.78}$ & ${\bf 77.75\pm 0.92}$\\
        MPCA  & $55.77\pm 1.08$ \\
        R-UMLDA & $46.62\pm 2.13$ \\
        \hline
        {\bf Probe C}& & & &\\
        HOOI & $36.67\pm 2.84$ & $36.73\pm 2.79$ & $36.70\pm 3.07$ & $36.87\pm 3.68$\\
        MPS & ${\bf 41.46\pm 0.64}$ & ${\bf 41.36\pm 0.64}$ & ${\bf 41.29\pm 0.63}$ & ${41.46\pm 0.59}$\\
        TTPCA & ${ 39.17\pm 0.90}$ & ${ 40.83\pm 0.41}$ & ${ 41.61\pm 1.02}$ & ${\bf 44.40\pm 1.54}$\\
        MPCA & $29.35\pm 2.29$ \\
        R-UMLDA & $20.87\pm 0.76$\\
        \hline
        {\bf Probe D}& & & &\\
        HOOI & $19.73\pm 0.91$ & $19.96\pm 1.15$ & $20.32\pm 0.93$ & $20.29\pm 1.11$\\
        MPS & ${\bf23.82\pm 0.42}$ & ${\bf 23.84\pm 0.43}$ & ${\bf 23.84\pm 0.45}$ & ${\bf23.84\pm 0.40}$\\
        TTPCA & ${ 21.92\pm 0.54}$ & ${ 22.14\pm 0.20}$ & ${ 22.84\pm 0.42}$ & ${ 21.92\pm 0.59}$\\
        MPCA & $21.11\pm 3.43$\\
        R-UMLDA & $7.88\pm 1.00$\\
        \hline
        {\bf Probe F}& & & &\\
        {HOOI} & ${\bf20.77\pm 0.92}$ & ${\bf20.71\pm 0.72}$ & $20.15\pm 0.65$ & $19.96\pm 0.67$\\
        {MPS} & ${ 20.50\pm 0.40}$ & ${ 20.52\pm 0.34}$ & ${\bf 20.50\pm 0.29}$ & ${\bf 20.56\pm 0.46}$\\
        TTPCA & ${ 14.78\pm 0.60}$ & ${ 14.74\pm 0.77}$ & ${ 15.29\pm 0.75}$ & ${ 15.40\pm 0.55}$\\
        MPCA & $17.12\pm 2.79$\\
        R-UMLDA & $9.67\pm 0.58$\\
        \hline
    \end{tabular}
\end{table*}
\begin{table*}[!htbp]
\normalsize
    \caption{Seven experiments in the USF GAIT dataset}
    \label{usfgprobes}
    \centering 
    \begin{tabular}{c|c|c|c|c|c|c|c}
         Probe set & A(GAL) & B(GBR) & C(GBL) & D(CAR) & E(CBR) & F(CAL) & G(CBL)\\
         \hline
         Size & 71&41&41&70&44&70&44\\
         \hline
         Differences&View&Shoe&Shoe, view&Surface&Surface, shoe&Surface, view&Surface, view, shoe\\
         \hline
    \end{tabular}
\end{table*}

In this experiment,  the core tensors remains very large even with a small threshold used, e.g., for $\epsilon=0.75$, the core size of each sample obtained by TTPCA/MPS and HOOI are $18\times 201 = 3618$ and $14\times 15\times 13 = 2730$, {\color{black}respectively, because of slowly decaying singular values, which make them too large for classification.} Therefore, we need to further reduce the sizes of core tensors before feeding them to classifiers for a better performance. In our experiment, we simply apply a further truncation to each core tensor by keeping the first few dimensions of each mode of the tensor. Intuitively, this can be done as we have already known that the space of each mode is orthogonal and ordered in such a way that the first dimension corresponds to the largest singular value, the second one corresponds to the second largest singular value and so on. Subsequently, we can independently truncate the dimension of each mode to a reasonably small value (which can be determined empirically) without changing significantly the meaning of the core tensors. It then gives rise to core tensors of smaller size that can be used directly for classification. More specifically, suppose that the core tensors obtained by MPS and HOOI have sizes $Q\times\De_1\times\De_2$ and $Q\times\De_1\times\De_2\times\De_3$, where $Q$ is the number $K$ ($L$) of training (test) samples, respectively. The core tensors are then truncated to be $Q\times\tilde{\De}_1\times\tilde{\De}_2$ and $Q\times\tilde{\De}_1\times\tilde{\De}_2\times\tilde{\De}_3$, respectively such that $\tilde{\De}_l<\De_l$ ($l=1,2,3$). Note that each $\tilde{\De}_l$ is chosen to be the same for both training and test core tensors. In regards to TTPCA, {\color{black} each core matrix is vectorized to have $\De_1\De_2$ features}, then PCA is applied.

{\color{black}Classification results for different threshold values $\epsilon$ is shown} in Table.~\ref{tablecEYFB} for TTPCA, MPS and HOOI using two different classifiers, i.e. KNN-1 and LDA. {\color{black}Results from MPCA and R-UMLDA is also included. The} core tensors obtained by MPS and HOOI are reduced to have sizes of $Q\times\tilde{\De}_1\times\tilde{\De}_2$ and $Q\times\tilde{\De}_1\times\tilde{\De}_2\times\tilde{\De}_3$, respectively such that $\tilde{\De}_1=\tilde{\De}_2=\De\in(10,11,12,13,14)$ and $\tilde{\De}_3=1$. Therefore, the reduced core tensors obtained by both methods have the same size for classification. With MPS and HOOI, each value of CSR in Table.~\ref{tablecEYFB} is computed by taking the average of the ones obtained from classifying different reduced core tensors due to different $\De$. In regards to TTPCA, for each $\epsilon$, a range of principal components $p=\{50,\ldots,70\}$ is used. We utilize $Q=\{70,75,80,85,90\}$ for MPCA, and the range $m_p=\{10,\ldots,20\}$ for R-UMLDA. The average CSR's are computed with TTPCA, MCPA and R-UMLDA according to their respective range of parameters in Table.~\ref{tablecEYFB}. We can see that the MPS gives rise to better results for all threshold values using different classifiers. More importantly, MPS with the smallest $\epsilon$ can produce the highest CSR. The LDA classifier gives rise to the best result, i.e. ${\bf 97.32\pm 0.89}$.

\subsubsection{BCI Jiaotong}
The BCIJ dataset consists of single trial recognition for BCI electroencephalogram (EEG) data involving left/right motor imagery (MI) movements. The dataset includes five subjects and the paradigm required subjects to control a cursor by imagining the movements of their right or left hand for 2 seconds with a 4 second break between trials. Subjects were required to sit and relax on a chair, looking at a computer monitor approximately 1m from the subject at eye level. For each subject, data was collected over two sessions with a 15 minute break in between. The first session contained 60 trials (30 trials for left, 30 trials for right) and were used for training. The second session consisted of 140 trials (70 trials for left, 70 trials for right). The EEG signals were sampled at 500Hz and preprocessed with a filter at 8-30Hz, hence for each subject the data consisted of a multidimensional tensor $channel\times time\times Q$. The common spatial patterns (CSP) algorithm \cite{1615701} is a popular method for BCI classification that works directly on this tensor,  and provides a baseline for the proposed and existing tensor-based methods.
For the tensor-based methods, we preprocess the data by transforming the tensor into the time-frequency domain using complex Mortlet wavelets with bandwidth parameter $f_b=6$Hz (CMOR6-1) to make classification easier \cite{ZHA07, PHA11}. The wavelet center frequency $f_c=1$Hz is chosen. Hence, the size of the concatenated tensors are $62\ channels\times 23\ frequency\ bins\times 50\ time\ frames\times Q$.

We perform the experiment for all subjects. After applying the feature extraction methods MPS and HOOI, the core tensors still have high dimension, so we need to further reduce their sizes before using them for classification. For instance, the reduced core sizes of MPS and HOOI are  chosen to be $Q\times 12\times\De$ and $Q\times 12\times\De\times 1$, where $\De\in(8,\ldots,14)$, respectively. With TTPCA, the principal components $p=\{10,50,100,150,200\}$, $Q=\{70,75,80,85,90\}$ for MPCA and $m_p=\{10,\ldots,20\}$ for R-UMLDA. With CSP, we average CSR for a range of spatial components $s_c=\{2,4,6,8,10\}$.

The LDA classifier is utilized and the results are shown in Table.~\ref{tablecBCI} for different threshold values of TTPCA, MPS and HOOI. The results of MPCA, R-UMLDA and CSP are also included. MPS outperforms the other methods for Subjects 1 and 2, and is comparable to HOOI in the results for Subject 5. CSP has the highest CSR for Subjects 3 and 4, followed by MPS or TTPCA, which demonstrates the proposed methods being effective at reducing tensors to relevant features, more precisely than current tensor-based methods.

\begin{figure}[htpb!]
    \centering
    \includegraphics[scale=3.3]{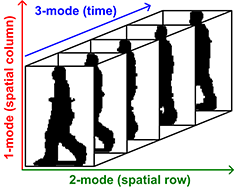}\\
    \caption{The gait silhouette sequence for a third-order tensor.}
    \label{gaitseqfig}
\end{figure}

\subsubsection{USF GAIT challenge}
The USFG database consists of 452 sequences from 74 subjects who walk in elliptical paths in front of a camera. There are three conditions for each subject: shoe type (two types), viewpoint (left or right), and the surface type (grass or concrete). A gallery set (training set) contains 71 subjects and there are seven types of experiments known as probe sets (test sets) that are designed for human identification. The capturing conditions for the probe sets is summarized in Table \ref{usfgprobes}, where G, C, A, B, L and R stand for grass surface, cement surface, shoe type A, shoe type B, left view and right view, respectively. The conditions in which the gallery set was captured is grass surface, shoe type A and right view (GAR). The subjects in the probe and gallery sets are unique and there are no common sequences between the gallery and probe sets. Each sequence is of size $128\times88$ and the time mode is 20, hence each gait sample is a third-order tensor of size $128\times88\times20$, as shown in Fig. \ref{gaitseqfig}. The gallery set contains 731 samples, therefore the training tensor is of size $128\times88\times20\times731$. The test set is of size $128\times88\times20\times P_s$, where $P_s$ is the sample size for the probe set that is used for a benchmark, refer to Table \ref{usfgprobes}. The difficulty of the classification task increases with the amount and and type of variables, e.g. Probe A only has the viewpoint, whereas Probe F has surface and viewpoint, which is more difficult. For the experiment we perform tensor object classification with Probes A, C, D and F (test sets).

The classification results based on using the LDA classifier is shown in Table~\ref{tablecGAIT}. The threshold $\epsilon$ still retains many features in the core tensors of MPS and HOOI. Therefore, further reduction of the core tensors is chosen to be $Q\times 20\times\De$ and $Q\times 20\times\De\times 1$, where $\De\in(8,\ldots,14)$, respectively.  The principal components for TTPCA is the range $p=\{150,200, 250,300\}$, $Q=\{70,75,80,85\}$ for MPCA and $m_p=\{10,\ldots,20\}$ for R-UMLDA. The proposed algorithms achieve the highest performance for Probes A, C, and D. MPS and HOOI are similar for the most difficult test set Probe F.

\begin{figure*}
\centering
  \mbox{\subfloat[COIL-100.]{\label{subfig:a} \includegraphics[height=6cm]{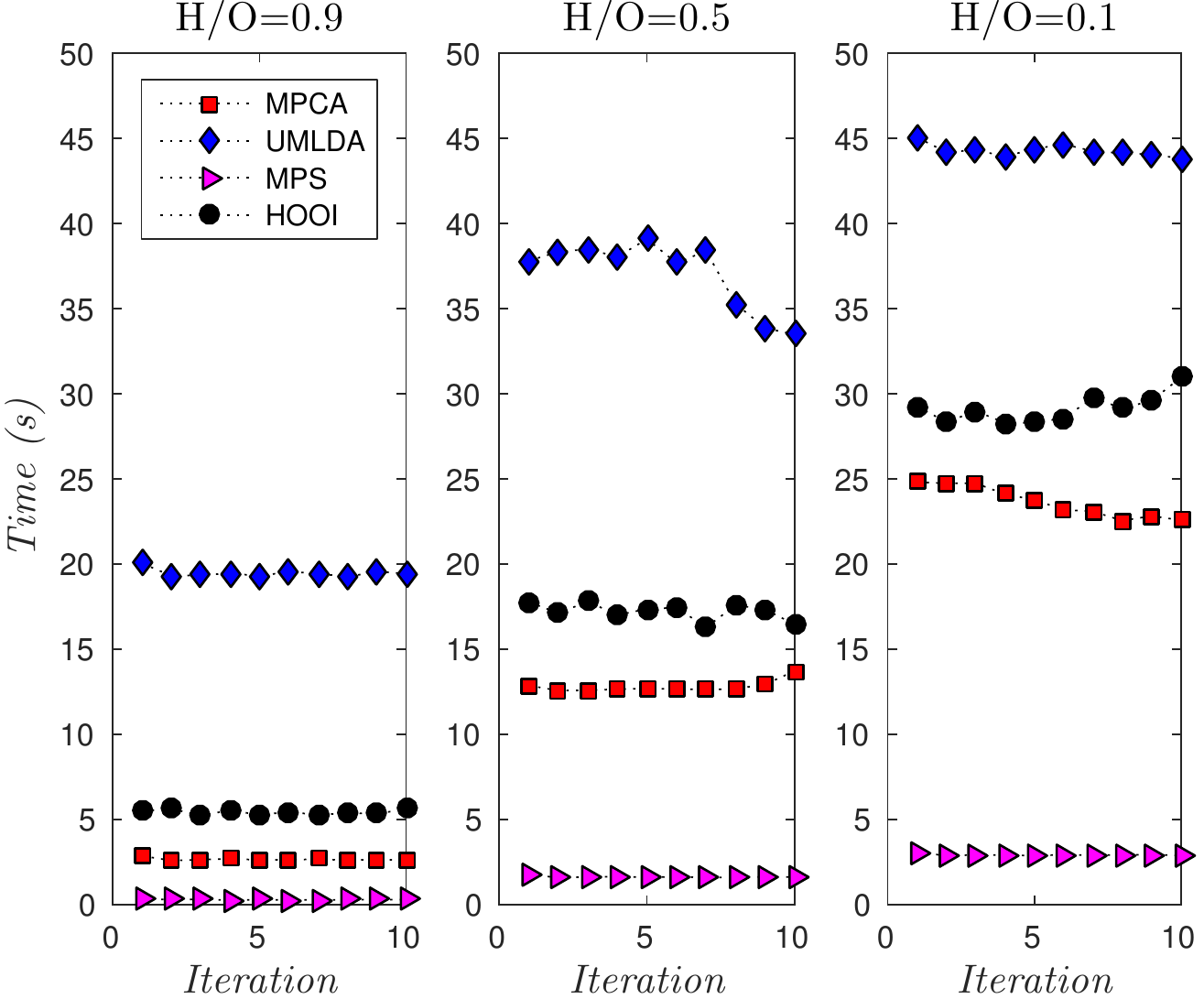}}}
  \mbox{\subfloat[EYFB.]{\label{subfig:b} \includegraphics[height=6cm]{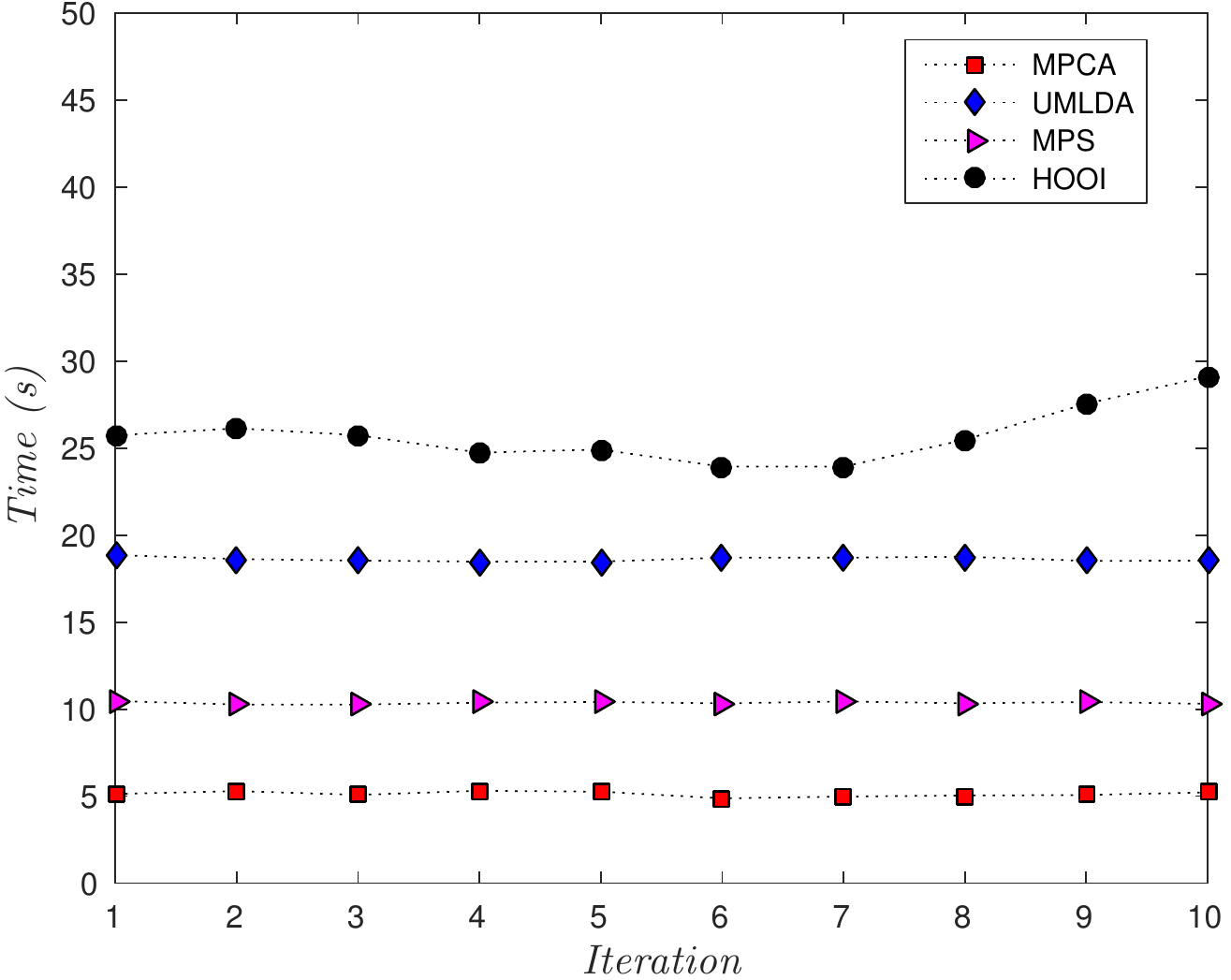}}}
    \mbox{\subfloat[BCI Subject 1.]{\label{subfig:b} \includegraphics[height=6cm]{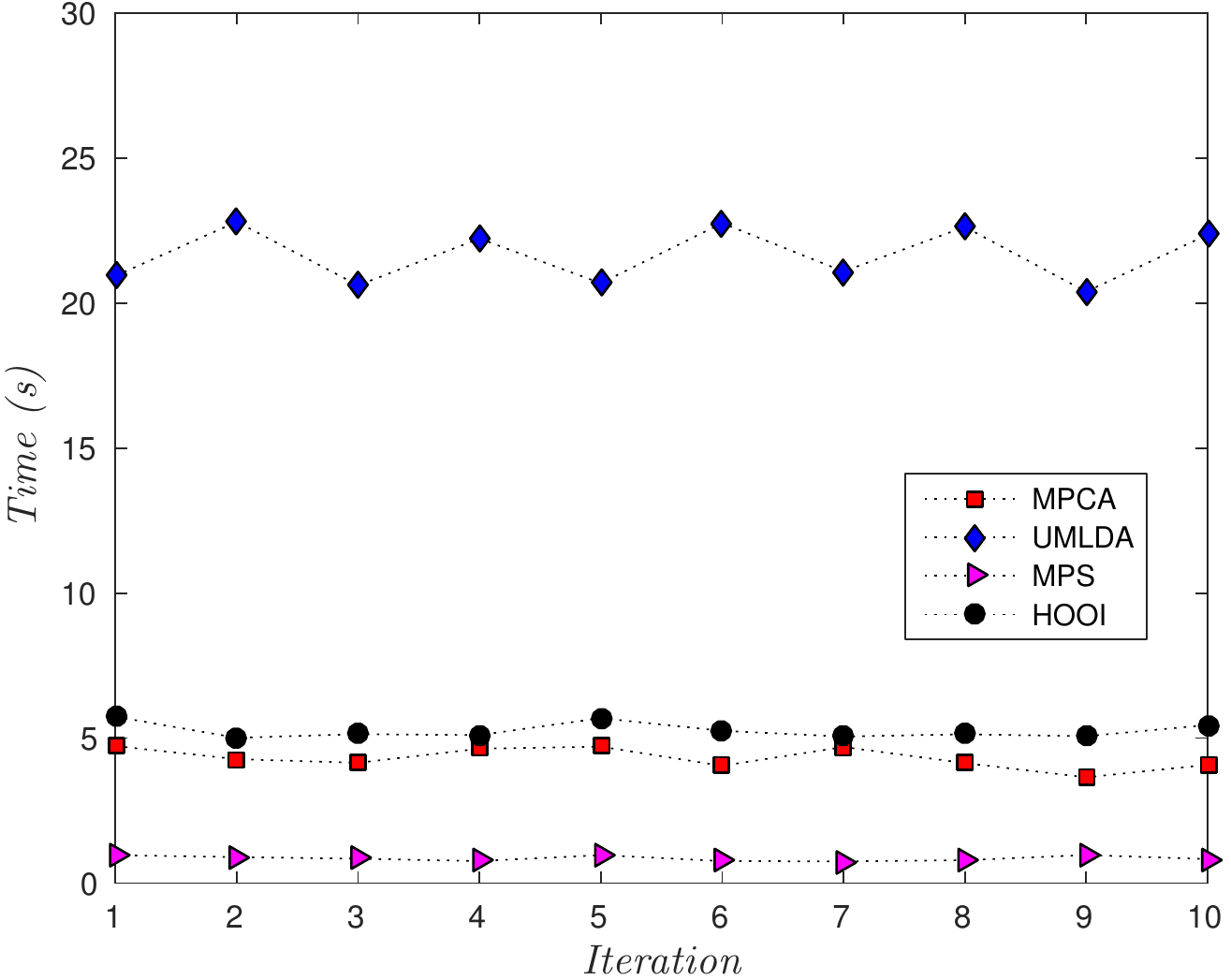}}}
      \mbox{\subfloat[GAIT Probe A.]{\label{subfig:b} \includegraphics[height=6cm]{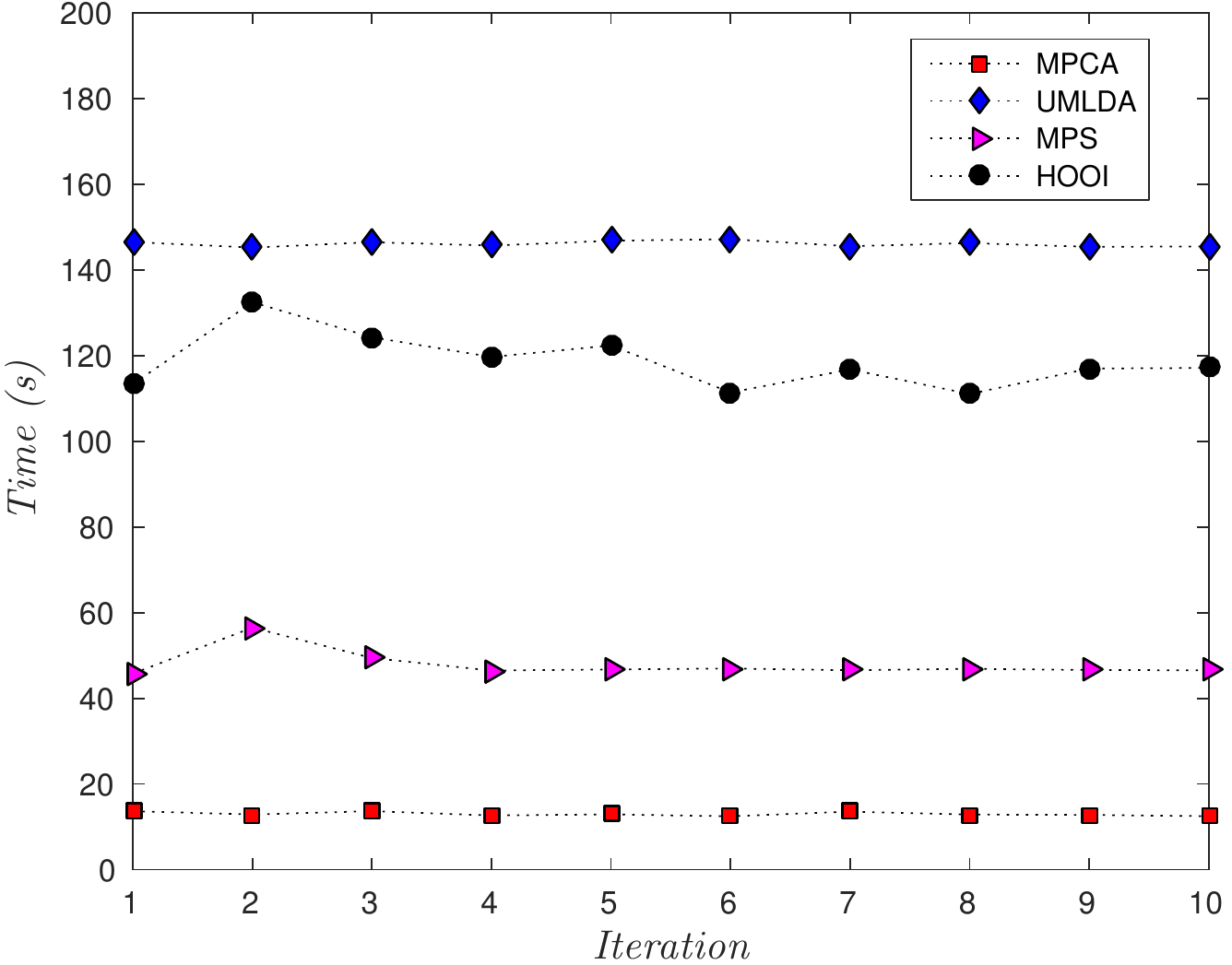}}}
      \caption{Training time of datasets for MPS, MPCA, HOOI and R-UMLDA.}\label{trtimefig}
\end{figure*}

\subsection{Training time benchmark}
An additional experiment on training time for MPS\footnote{TTPCA would be equivalent in this experiment.}, HOOI, MPCA and R-UMLDA is provided to understand the computational complexity of the algorithms.
For the COIL-100 dataset, we measure the elapsed training time for the training tensor of size $32\times32\times3\times K$ ($K=720,3600,6480$) for H/O$=\{0.9,0.5,0.1\}$, according to 10 random partitions of train/test data (\textit{iterations}). MPCA, HOOI and R-UMLDA reduces the tensor to 32 features, and MPS to 36 ({\color{black}due to a fixed dimension} $\Delta^2$). In Fig. \ref{trtimefig}a, we can see that as the number of training images increases, the MPS algorithms computational time only slightly increases, while MCPA and HOOI increases gradually, with UMLDA having the slowest performance overall.

The EYFB benchmark reduces the training tensor features to 36 (for MPS), 32 (MPCA and HOOI), and 16 (UMLDA, since the elapsed time for 32 features is too long). For this case, Fig. \ref{trtimefig}b demonstrates that MPCA provides the fastest computation time due to its advantage with small sample sizes (SSS). MPS performs the next best, followed by HOOI, then UMLDA with the slowest performance.

The BCI experiment involves reducing the training tensor to 36 (MPS) or 32 (MPS, HOOI and UMLDA) features and the elapsed time is shown for Subject 1 in Fig. \ref{trtimefig}c. For this case MPS performs the quickest compared to the other algorithms, with UMLDA again performing the slowest.

Lastly, the USFG benchmark tests Probe A by reducing the MPS training tensor to 36 features, MPCA and HOOI to 32 features, and UMLDA to 16 features. Fig. \ref{trtimefig}d shows that MPCA provides the quickest time to extract the features, followed by MPS, HOOI and lastly UMLDA.

\section{Conclusion}\label{Conclusions}
In this paper, {\color{black}a rigorous analysis of MPS and Tucker decomposition proves the efficiency of MPS in terms of retaining relevant correlations and features, which can be used directly for tensor object classification. Subsequently, two new approaches to tensor dimensionality reduction based on compressing tensors to matrices are proposed. One method reduces a tensor to a matrix, which then utilizes PCA. And the other is a new multidimensional analogue to PCA known as MPS. Furthermore, a comprehensive discussion on the practical implementation of the MPS-based approach is provided, which emphasizes tensor mode permutation, tensor bond dimension control, and core matrix positioning.
Numerical simulations demonstrates the efficiency of the MPS-based algorithms against other popular tensor algorithms for dimensionality reduction and tensor object classification.

For the future outlook, we plan to explore this approach to many other problems in multilinear data compression and tensor super-resolution.}

\bibliographystyle{IEEEtran}
\bibliography{ref}
\vfill
\end{document}